\documentclass[10pt,twocolumn,letterpaper]{article}

\usepackage[pagenumbers]{cvpr} 

\usepackage{colortbl}
\usepackage{diagbox}
\usepackage{multirow}
\usepackage{booktabs}

\definecolor{cvprblue}{rgb}{0.21,0.49,0.74}
\usepackage[pagebackref,breaklinks,colorlinks,allcolors=cvprblue]{hyperref}

\def\paperID{727} 
\def\confName{CVPR}
\def\confYear{2025}

\title{MMRL: Multi-Modal Representation Learning for Vision-Language Models}

\author{Yuncheng Guo$^1$, Xiaodong Gu$^2$\thanks{Corresponding author}\\
Department of Electronic Engineering, Fudan University, Shanghai 200438, China\\
{\tt\small $^1$23210720033@m.fudan.edu.cn, $^2$xdgu@fudan.edu.cn}
}

\begin{document}
\maketitle
\begin{abstract} 
Large-scale pre-trained \textbf{V}ision-\textbf{L}anguage \textbf{M}odels (VLMs) have become essential for transfer learning across diverse tasks. However, adapting these models with limited few-shot data often leads to overfitting, diminishing their performance on new tasks. To tackle this issue, we propose a novel \textbf{M}ulti-\textbf{M}odal \textbf{R}epresentation \textbf{L}earning (MMRL) framework that introduces a shared, learnable, and modality-agnostic representation space. MMRL projects the space tokens to text and image representation tokens, facilitating more effective multi-modal interactions. Unlike previous approaches that solely optimize class token features, MMRL integrates representation tokens at higher layers of the encoders—where dataset-specific features are more prominent—while preserving generalized knowledge in the lower layers. During training, both representation and class features are optimized, with trainable projection layer applied to the representation tokens, whereas the class token projection layer remains frozen to retain pre-trained knowledge. Furthermore, a regularization term is introduced to align the class features and text features with the zero-shot features from the frozen VLM, thereby safeguarding the model's generalization capacity. For inference, a decoupling strategy is employed, wherein both representation and class features are utilized for base classes, while only the class features, which retain more generalized knowledge, are used for new tasks. Extensive experiments across 15 datasets demonstrate that MMRL outperforms state-of-the-art methods, achieving a balanced trade-off between task-specific adaptation and generalization. Code is available at \href{https://github.com/yunncheng/MMRL}{https://github.com/yunncheng/MMRL}.
\end{abstract}
    
\section{Introduction}
\label{sec:intro}
Vision-Language Models (VLMs) \cite{clip, align, flamingo, filip, kosmos1, kosmos2, vila}, such as CLIP \cite{clip}, have gained significant attention for their ability to leverage the rich, complementary information inherent in both textual and visual modalities. By constructing distinct encoders for images and text, and employing contrastive learning \cite{contrastive_learning} on over 400 million image-text pairs, CLIP effectively captures complex visual-text relationships, demonstrating strong performance across various downstream tasks, such as medical image analysis \cite{clip_medical1, clip_medical2, clip_medical3}, image and video captioning \cite{clip_captioning1, clip_captioning2, clip_captioning3}, and visual question answering \cite{clip_answering1, clip_answering2, clip_answering3}. Despite their versatility, VLMs encounter limitations in adapting to new tasks, as fine-tuning their large-scale architectures demands considerable computational resources.


\begin{figure}[tb]
\centering
\setlength{\abovecaptionskip}{0.2cm}   
  \includegraphics[width=0.9\linewidth]{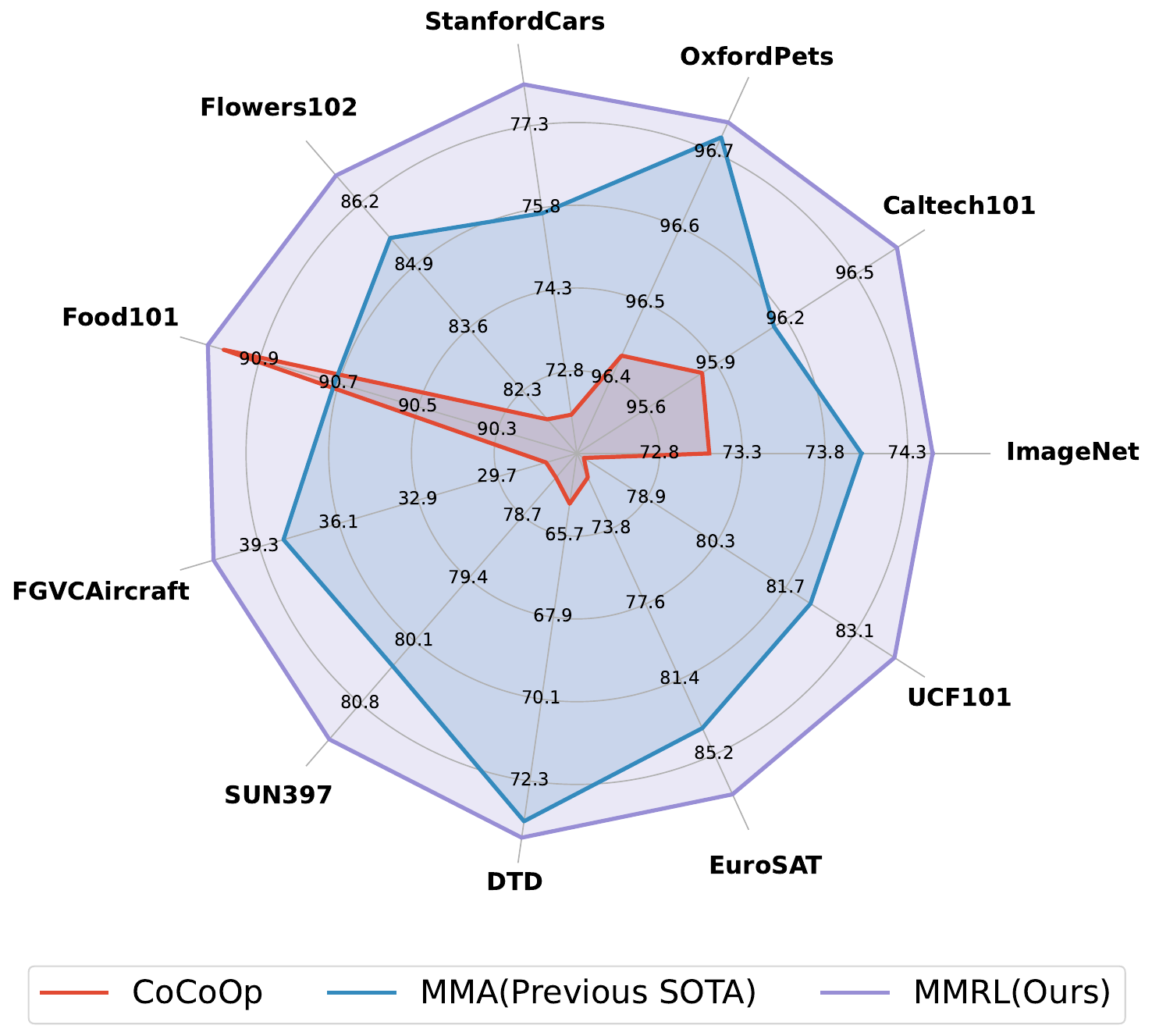}
  \caption{Comprehensive comparison of the harmonic mean performance between the previous sota method MMA and our proposed MMRL across 11 diverse datasets for base-to-novel generalization. Our method achieves the best on all datasets.}
  \label{radar}
\vspace{-0.5cm}
\end{figure}

To facilitate efficient adaptation of VLMs, strategies such as prompt engineering and ensembling \cite{clip} have shown potential. Specifically, prompt engineering involves crafting dataset-specific prompts, such as ``A photo of a [CLASS], a type of pet.'' for the OxfordPets \cite{oxford_pets} dataset. Alternatively, ensembling can integrate multiple zero-shot classifiers by varying context prompts, \eg, ``A photo of a big [CLASS].'' and ``A photo of a small [CLASS].''. Nonetheless, manual prompt design is time-consuming and requires substantial expertise, yet it does not guarantee the discovery of optimal prompts. To address this limitation, CoOp \cite{coop} introduces prompt learning\cite{prompt_tuning} where prompts are modeled as continuous learnable vectors, optimized during training while keeping VLM parameters fixed, thereby enabling efficient dataset adaptation. Recently MaPLe \cite{maple} has identified that prompt learning solely within the text modality may be sub-optimal. In response, it proposes a multi-modal prompt learning approach, embedding deep prompts into the lower layers of both VLM encoders via a coupling function to enhance alignment between visual and textual representations.

In addition to prompt learning, adapter-style learning methods offer a different adaptation pathway: rather than modifying input prompts, lightweight modules (\eg, multi-layer perceptrons, MLPs) are integrated within VLMs to adjust extracted features for downstream datasets. CLIP-Adapter \cite{clip-adapter} exemplifies this approach by maintaining the frozen VLM while fine-tuning features via an MLP adapter added to the image encoder, which incorporates residual connections for feature fusion. Similar to MaPLe, MMA \cite{mma} proposes a multimodal adapter that refines the alignment between text and vision representations by aggregating features from diverse branches into a unified feature space, allowing gradient flow across branches. Notably, MMA reveals that different layers within VLM encoders capture varying characteristics: higher layers encode discriminative, dataset-specific information, while lower layers retain more generalizable features.

However, the current multimodal deep prompt learning method \cite{maple}, which applies prompt concatenation at shallow layers, may compromise generalizable knowledge. This approach map visual prompts from text prompts, incorporating visual information via gradient propagation but ultimately remaining text-centric, with updates focused mainly on text prompts. Moreover, both prompt learning and adapter-style methods solely optimize class token features using task-specific objectives, such as cross-entropy loss. As a result, these methods are vulnerable to overfitting to specific data distributions or task categories when training data is scarce (\eg, few-shot setting), leading to a decline in the inherent generalization and zero-shot learning capabilities of VLMs.

To address these challenges, we propose a novel multi-modal representation learning framework that distinguishes itself from conventional prompt learning and adapter-style methods. Specifically, we introduce a shared, learnable representation space that is independent of any modality within the higher layers of the encoder. This space serves as a bridge for multimodal interaction, mapping tokens from this space to both image and text representation tokens, which are then concatenated with the original encoder tokens to enable effective multimodal interaction. Our representation tokens are designed to learn dataset-specific knowledge from downstream tasks while the original classification token is regularized to retain a significant amount of generalizable knowledge. MMRL offers three key advantages: (1) an unbiased shared representation space that promotes balanced multimodal learning; (2) preservation of original VLM generalization by avoiding prompt integration at shallow encoder layers; and (3) Unlike prompt learning or adapter-style methods that refine only the class token features through learnable prompts or adapters, our approach supports decoupled inference across classes. During training, we prioritize optimizing representation token features, with their projection layer trainable, while that of the original class token remain fixed. To further preserve the generalizability of the class token, a regularization term aligns its features with the zero-shot features from the frozen VLM. For inference, we utilize both representation and class token features for base classes, while for unseen classes or new datasets, only the class token features are employed.

Our main contributions are summarized as follows:
\begin{itemize}
    \item We introduce the \textbf{M}ulti-\textbf{M}odal \textbf{R}epresentation \textbf{L}earning (MMRL) framework, which incorporates a shared, unbiased, learnable space that bridges image and text modalities, facilitating multimodal interaction at the high layers of the original encoder.
    \item A decoupling strategy preserves VLM generalization by adapting representation tokens for downstream tasks while regularizing the original class token for new tasks.
    \item Extensive experiments demonstrate that MMRL substantially improves downstream adaptation and generalization, achieving superior performance over baselines.
\end{itemize}


\begin{figure*}
\centering
\setlength{\abovecaptionskip}{0.2cm}   
  \includegraphics[width=0.9\linewidth]{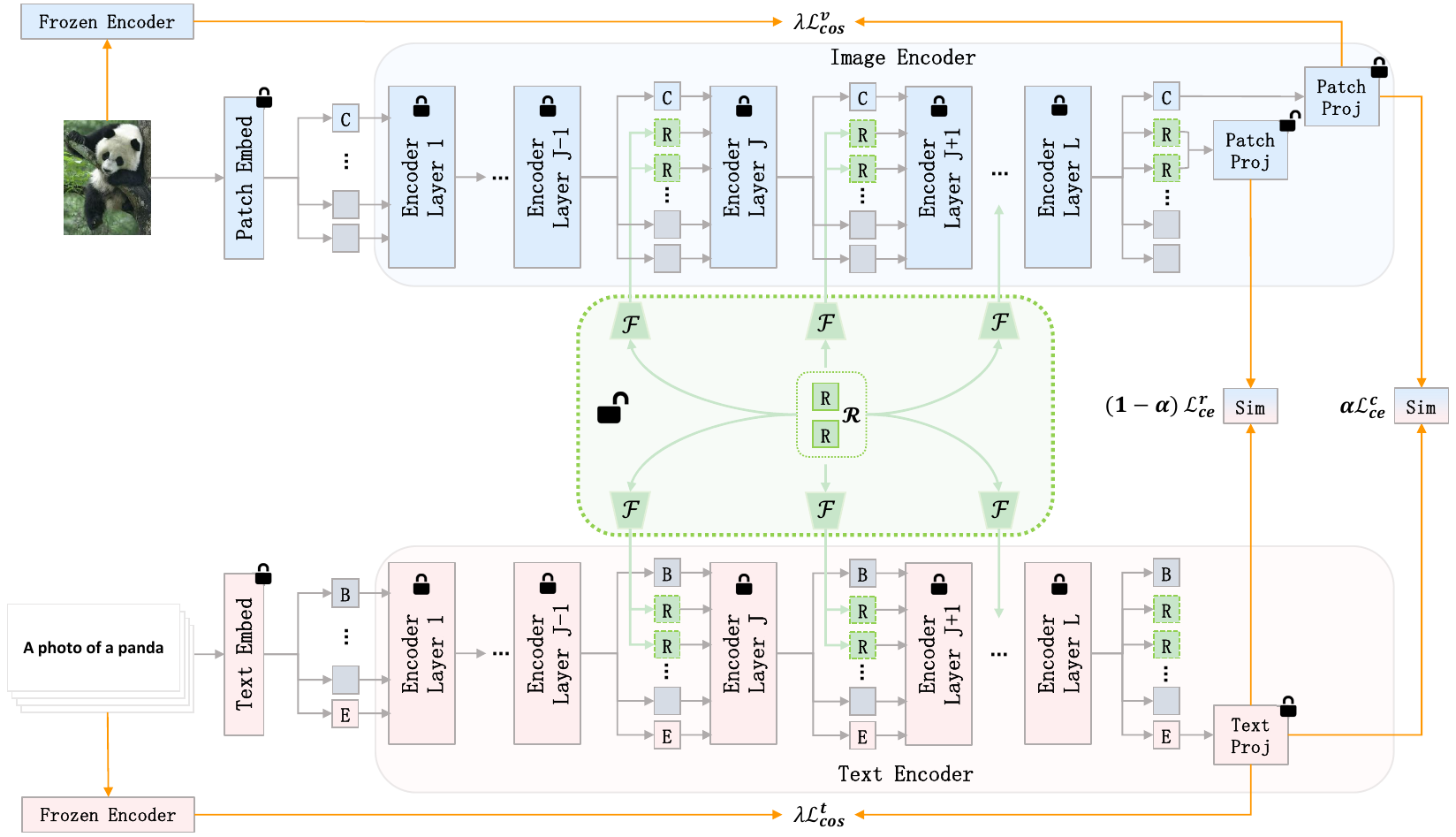}
  \caption{MMRL training framework. Here, `C' denotes the class token, `B' the BOT token, `E' the EOT token, $\mathcal{R}$ our representation space, and `R' the representation token. Only the representation space $\mathcal{R}$, mapping function $\mathcal{F}$, and the patch projection layer for the representation tokens are optimized, while the entire pre-trained CLIP model remains frozen. To preserve generalization knowledge, we integrate representation tokens in both encoders starting from layer $J$.}
  \label{framework1}
\vspace{-0.5cm}
\end{figure*}

\section{Related Work}
\label{sec:related work}
\subsection{Vision-Language Models}
Vision-language models (VLMs) have emerged as powerful tools for capturing rich multimodal representations, standing apart from traditional models that rely exclusively on visual or textual supervision. Recent advances in VLMs, such as CLIP \cite{clip}, ALIGN \cite{align}, FILIP \cite{filip}, KOSMOS \cite{kosmos1, kosmos2}, and VILA \cite{vila}, have demonstrated remarkable performance across a variety of tasks. These models typically learn joint image-language representations through self-supervised learning, leveraging large-scale architectures and massive collections of image-text pairs. For instance, CLIP is trained on a collection of 400 million image-text pairs, while ALIGN leverages an impressive 1.8 billion pairs. Although these pre-trained models excel at learning generalized representations, efficiently adapting them to specific downstream tasks remains a challenge.

\subsection{Efficient Transfer Learning}
Prompt learning methods have proven effective for adapting VLMs. CoOp \cite{coop} pioneers prompt learning \cite{prompt_tuning, prefix_tuning, p_tuning} by replacing fixed templates with learnable continuous vectors, enhancing flexibility but compromising CLIP's zero-shot and generalization capabilities. To address this, CoCoOp \cite{cocoop} incorporates visual cues to generate instance-specific prompts, improving generalization to class distribution shifts, while ProDA \cite{proda} learns prompt distributions to enhance adaptability. PLOT \cite{plot} uses optimal transport to align the vision and text modalities. KgCoOp \cite{kgcoop} retains general textual knowledge by minimizing divergence between learned and crafted prompts. ProGrad \cite{prograd} selectively updates gradients aligned with general knowledge, and RPO \cite{rpo} mitigates internal representation shifts using masked attention. Moving beyond text-focused approaches, MaPLe \cite{maple} integrates visual prompts mapped from text prompts through a coupling function, fostering cross-modal synergy. ProVP \cite{provp} employs single-modal visual prompts with contrastive feature re-formation to align prompted visual features with CLIP's distribution. PromptSRC \cite{promptsrc} employs a self-regularization strategy to mitigate overfitting, while MetaPrompt \cite{metaprompt} applies a meta-learning-based prompt tuning algorithm that encourages task-specific prompts to generalize across various domains or classes. TCP \cite{tcp} adapts textual knowledge into class-aware tokens, enhancing generalization capabilities.

Adapter-style learning methods represent another efficient pathway for VLM adaptation. CLIP-Adapter \cite{clip-adapter} uses lightweight adapters, implemented as two-layer MLPs, to refine CLIP's feature representations through cross-entropy optimization. Building on this, Tip-Adapter \cite{tip-adapter} caches training features to facilitate efficient similarity calculations between test and training features. However, both methods process image and text representations independently before prediction. Addressing this separation, MMA \cite{mma} integrates features across branches into a shared space, allowing for cross-branch gradient flow and enhanced coherence between modalities.

In addition to the aforementioned methods, several approaches \cite{coprompt, hpt, argue, promptkd} leverage large language models (LLMs) such as GPT-3 \cite{gpt3} for text augmentation or apply distillation over the entire dataset to improve performance. However, the increased computational requirements associated with these methods may place them beyond the intended scope of efficient transfer learning.
\section{Method}
\label{sec:method}
Our approach, in line with previous methods, builds upon a pre-trained VLM, CLIP \cite{clip}. In this section, we detail the construction of our MMRL framework and the implementation specifics.

\subsection{Preliminary}
We begin by defining the notations used in our approach. CLIP comprises two encoders: an image encoder $\mathcal{V}$ and a text encoder $\mathcal{W}$.

\noindent \textbf{Image Encoding:} The image encoder $\mathcal{V}$ consists of $L$ transformer \cite{transformer} layers, denoted $\{\mathcal{V}_i\}_{i=1}^{L}$. Given an input image \( x \in \mathbb{R}^{H \times W \times 3} \), it is divided into \( M \) fixed-size patches, each projected into a patch embedding, resulting in \( E_0 \in \mathbb{R}^{M \times d_v} \), where $M$ represents the number of patches and $d_v$ the embedding dimension. The initial patch embeddings $E_0$ are combined with a learnable class token $c_0$ and positional encodings, forming the input sequence for the transformer layers. Each layer processes this sequence as
\begin{equation}
    [c_i, E_i] = \mathcal{V}_i([c_{i-1}, E_{i-1}]) \quad
    i = 1, 2, \ldots, L
    \nonumber
\end{equation}
After passing through all transformer layers, a patch projection layer, $P_v^c$, projects the output of the class token, $c_L$, into a shared V-L latent space,
\begin{equation}
    f = P_v^c(c_L)
    \nonumber
\end{equation}
where $f \in \mathbb{R}^{d}$.

\noindent \textbf{Text Encoding:} For an input text, \eg, ``A photo of a [CLASS].", it is tokenized and converted into embeddings $T_0 \in \mathbb{R}^{N \times d_t}$, where $N$ is the token length and $d_t$ the embedding dimension. Beginning-of-text (BOT) and end-of-text (EOT) tokens, denoted $b_0$ and $e_0$, mark the sequence boundaries. These token embeddings, with positional encodings, are passed through the text encoder's $L$ transformer layers, $\{\mathcal{W}_i\}_{i=1}^{L}$, as follows,
\begin{equation} 
    [b_i, T_i, e_i] = \mathcal{W}_i([b_{i-1}, T_{i-1}, e_{i-1}]) \quad i = 1, \ldots, L 
    \nonumber
\end{equation} 
After the final layer, the output of the EOT token, $e_L$, is projected into the shared V-L space using $P_t$,
\begin{equation} 
    w = P_{t}(e_{L}) \nonumber
\end{equation} 
where $w \in \mathbb{R}^{d}$.

\noindent \textbf{Classification with CLIP:} With the image feature $f$ and text features $\{w_c\}_{c=1}^C$ for $C$ classes, CLIP calculates the cosine similarity between $f$ and each $w_c$,
\begin{equation} 
    \text{sim}(f, w_c) = \frac{f \cdot w_c}{|f| |w_c|}, \nonumber 
\end{equation} 
where $|\cdot|$ represents the $L_2$ norm. Class probabilities are then computed using the softmax function,
\begin{equation} 
    p(y = c \mid f) = \frac{\exp(\text{sim}(f, w_c) / \tau)}{\sum_{i=1}^{C} \exp(\text{sim}(f, w_i) / \tau)} \nonumber 
\end{equation} 
where $\tau$ is a temperature parameter. The final predicted class is selected as the one with the highest probability score.


\begin{figure*}[tb]
\setlength{\abovecaptionskip}{0.2cm}   
\setlength{\belowcaptionskip}{-0.4cm}   
\centering
\setlength{\belowcaptionskip}{-0.39cm}   
  \includegraphics[width=0.7\linewidth]{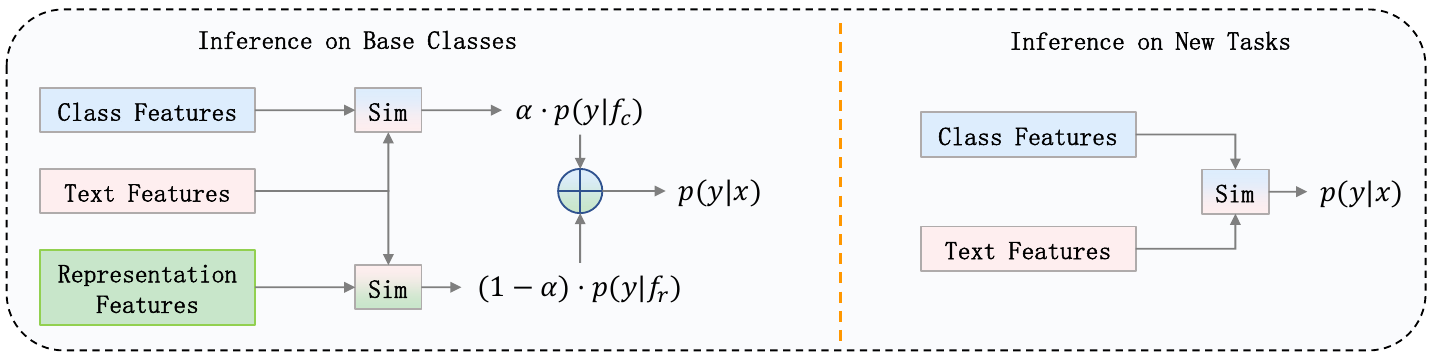}
  \caption{MMRL inference process, where different tasks utilize distinct features.}
  \label{framework2}
\end{figure*}

\subsection{Multi-Modal Representation Learning (MMRL)} Our proposed MMRL aims to address the challenges of adapting pre-trained VLMs using few-shot data while maintaining generalization to new tasks. The training and inference frameworks of MMRL are shown in \cref{framework1} and \cref{framework2}, respectively. In the following, we describe the specifics of the methodology.

\subsubsection{Learnable Representation Space}
MMRL establishes a shared, learnable representation space $\mathcal{R}$ to facilitate multimodal interactions, initialized through sampling from a Gaussian distribution. Using a learnable mapping function $\mathcal{F}(\cdot)$, implemented as a linear layer, we project the tokens $R \in \mathbb{R}^{K \times d_r}$ in this space—where $K$ is the number of tokens and $d_r$ is the dimension of the representation space—into both visual and textual modalities,
\begin{align}
    R^v = \{R_i^v\}_{i=J-1}^{L-1} \quad & R_i^v = \mathcal{F}_i^v(R) \nonumber \\ 
    R^t = \{R_i^t\}_{i=J-1}^{L-1} \quad & R_i^t = \mathcal{F}_i^t(R) \nonumber 
\end{align}
where $R_i^v \in \mathbb{R}^{K \times d_v}$ and $R_i^t \in \mathbb{R}^{K \times d_t}$ represent the representation tokens for visual and textual modalities, respectively, in the $(i+1)$-th transformer layer. The index $J$ indicates the starting layer from which these representation tokens are integrated into the encoders.

\subsubsection{Integration into Higher Encoder Layers}
To preserve the generalized knowledge in the lower layers of the pre-trained CLIP model, the representation tokens $\mathcal{R}^v$ and $\mathcal{R}^t$ are integrated into the higher layers of the image encoder $\mathcal{V}$ and the text encoder $\mathcal{W}$, beginning from the $J$-th layer.

For the image encoder $\mathcal{V}$,
\begin{align}
    [c_i, E_i] &= \mathcal{V}_i([c_{i-1}, E_{i-1}]) \quad i = 1, \ldots, J-1 \nonumber \\  
    [c_i, \_, E_i] &= \mathcal{V}_i([c_{i-1}, R_{i-1}^v, E_{i-1}]) \quad i = J, \ldots, L - 1 \nonumber \\
    [c_i, R_i^v, E_i] &= \mathcal{V}_i([c_{i-1}, R_{i-1}^v, E_{i-1}]) \quad i = L \nonumber
\end{align}

For the text encoder $\mathcal{W}$, while previous prompt learning \cite{maple} involves replacing parts of $T_i$ to incorporate deep prompts, we retain the entire $T_i$ and insert $R_i^t$ before it, aiming to preserve the original textual information,
\begin{align}
    [b_i, T_i, e_i] &= \mathcal{W}_i([b_{i-1}, T_{i-1}, e_{i-1}]) \quad i = 1, \ldots, J-1 \nonumber \\  
    [b_i, \_, T_i, e_i] &= \mathcal{W}_i([b_{i-1}, R_{i-1}^t, T_{i-1}, e_{i-1}]) \nonumber \\
    &\hspace{4cm} i = J, \ldots, L-1 \nonumber \\
    [b_i, R_i^t, T_i, e_i] &= \mathcal{W}_i([b_{i-1}, R_{i-1}^t, T_{i-1}, e_{i-1}]) \quad i = L \nonumber
\end{align}
Note that due to the autoregressive nature of the text encoder, we adjust the attention mask matrix to accommodate the increased embedding length.

\subsubsection{Representation Learning}
Representation learning is designed to leverage representation tokens for dataset-specific adaptation, while the class token preserves the pre-trained knowledge of the original CLIP. Through a set of strategies aimed at retaining generalization during both training and inference, MMRL enables flexible inference for different tasks, as detailed below.

\begin{itemize} 

\item \textbf{Training Phase:} We optimize the features of both the representation tokens and the original class token, with the primary focus on representation features to preserve pre-trained knowledge. Specifically, the projection layer for the representation tokens is trainable, while that for the class token remains fixed. For the image encoder $\mathcal{V}$, after passing through $L$ transformer layers, we obtain the output $c_L \in \mathbb{R}^{d_v}$ for the class token and $R_L^v \in \mathbb{R}^{K \times d_v}$ for the $K$ representation tokens. The final output of the representation tokens, $r_L$, is derived by averaging across the $K$ tokens,
\begin{equation}
    r_L = \text{Mean}(R_L^v) \nonumber
\end{equation}
where $r_L \in \mathbb{R}^{d_v}$. We then apply the patch projection layers to map the outputs of both the class and representation tokens into the common V-L latent space, yielding the class features $f_c$ and representation features $f_r$.
\begin{equation}
    f_c = P_v^c(c_L) \quad f_r = P_v^r(r_L) \nonumber
\end{equation}
Here, $P_v^c$ is the original, frozen patch projection layer of CLIP for class features, while $P_v^r$ for representation features is trainable.

For the text encoder $\mathcal{W}$, following the sequential nature of text, we map the EOT token $e_L$—as in the original CLIP model—after processing through $L$ transformer layers into the common V-L space, yielding the text features.
\begin{equation} 
    w = P_{t}(e_{L}) \nonumber
\end{equation}
With the image features $f_c$, $f_r$, and the text classifiers $\{w_c\}_{c=1}^C$ for $C$ classes, we apply cross-entropy loss to separately optimize the class and representation features,
\begin{align}
\setlength\abovedisplayskip{3pt}
\setlength\belowdisplayskip{3pt}
    \mathcal{L}_{ce}^c &= -\sum_c^C y_c \log p(y = c \mid f_c) \nonumber \\
    \mathcal{L}_{ce}^r &= -\sum_c^C y_c \log p(y = c \mid f_r) \nonumber
\end{align}
where $y_c = 1$ if the image $x$ belongs to class $c$, and $y_c = 0$ otherwise. To further preserve the generalization of class features, we maximize the cosine similarity between $(f_c, w)$ and the frozen CLIP features $(f_0, w_0)$, explicitly guiding the training trajectory,
\begin{equation}
    \mathcal{L}_{cos}^v = 1 - \frac{f_c \cdot f_0}{|f_c| |f_0|} \quad \mathcal{L}_{cos}^t = 1 - \frac{1}{C}\sum_c^C \frac{w^c \cdot w_0^c}{|w^c| |w_0^c|}, \nonumber
\end{equation}
The final MMRL loss function is
\begin{equation}
    \mathcal{L}_{MMRL} = \alpha \mathcal{L}_{ce}^c + (1 - \alpha) \mathcal{L}_{ce}^r + \lambda (\mathcal{L}_{cos}^v + \mathcal{L}_{cos}^t)  \nonumber
\end{equation}
where $\alpha$ controls the balance between the features, and $\lambda$ is the penalty coefficient.

\item \textbf{Testing on Base Classes:} 
For in-distribution classes seen during training, we combine the dataset-specific representation features with the class features that preserve generalizability. The probability of an in-distribution test sample $x$ belonging to the $c$-th class is
\begin{equation}
    p(y = c \mid x) = \alpha \cdot p(y = c \mid f_c) + (1-\alpha) \cdot p(y = c \mid f_r) \nonumber
\end{equation}
where $f_c$ and $f_r$ are features extracted from the class token and representation tokens, respectively.

\item  \textbf{Testing on Novel Classes:}
For classes unseen during training or for new datasets, we rely solely on the class tokens, which retain generalized knowledge.
\begin{equation}
    p(y = c \mid x) = p(y = c \mid f_c) \nonumber
\end{equation}
\end{itemize}


\begin{table*}[t]
\centering
\setlength{\abovecaptionskip}{0.15cm}   
\caption{Comparison of MMRL with previous state-of-the-art methods on base-to-novel generalization across 11 datasets. Bold values indicate the best results. MMRL consistently enhances base class performance without compromising generalization.}
\label{base_to_novel}
\renewcommand\arraystretch{1.06}
\setlength{\tabcolsep}{2.87mm}{
\resizebox{0.85\textwidth}{!}{
    \begin{tabular}{@{}r|ccc|ccc|ccc|ccc@{}}
    \toprule
    \multirow{2}{*}{Method} &
      \multicolumn{3}{c|}{Average} &
      \multicolumn{3}{c|}{ImageNet} &
      \multicolumn{3}{c|}{Caltech101} &
      \multicolumn{3}{c}{OxfordPets} \\
     &
      Base &
      Novel &
      HM &
      Base &
      Novel &
      HM &
      Base &
      Novel &
      HM &
      Base &
      Novel &
      HM \\ \midrule
    $\text{CLIP}_{\text{ (ICML2021)}}$ &
      69.34 &
      74.22 &
      71.70 &
      72.43 &
      68.14 &
      70.22 &
      96.84 &
      94.00 &
      95.40 &
      91.17 &
      97.26 &
      94.12 \\
    $\text{CoOp}_{\text{ (IJCV2022)}}$ &
      82.69 &
      63.22 &
      71.66 &
      76.47 &
      67.88 &
      71.92 &
      98.00 &
      89.81 &
      93.73 &
      93.67 &
      95.29 &
      94.47 \\
    $\text{CoOpOp}_{\text{ (CVPR2022)}}$ &
      80.47 &
      71.69 &
      75.83 &
      75.98 &
      70.43 &
      73.10 &
      97.96 &
      93.81 &
      95.84 &
      95.20 &
      97.69 &
      96.43 \\
    $\text{ProDA}_{\text{ (CVPR2022)}}$ &
      81.56 &
      72.30 &
      76.65 &
      75.40 &
      70.23 &
      72.72 &
      98.27 &
      93.23 &
      95.68 &
      95.43 &
      97.83 &
      96.62 \\
    $\text{KgCoOp}_{\text{ (CVPR2023)}}$ &
      80.73 &
      73.60 &
      77.00 &
      75.83 &
      69.96 &
      72.78 &
      97.72 &
      94.39 &
      96.03 &
      94.65 &
      97.76 &
      96.18 \\
    $\text{MaPLe}_{\text{ (CVPR2023)}}$ &
      82.28 &
      75.14 &
      78.55 &
      76.66 &
      70.54 &
      73.47 &
      97.74 &
      94.36 &
      96.02 &
      95.43 &
      97.76 &
      96.58 \\
    $\text{PromptSRC}_{\text{ (ICCV2023)}}$ &
      84.26 &
      76.10 &
      79.97 &
      77.60 &
      70.73 &
      74.01 &
      98.10 &
      94.03 &
      96.02 &
      95.33 &
      97.30 &
      96.30 \\
    $\text{ProVP}_{\text{ (IJCV2024)}}$ &
      85.20 &
      73.22 &
      78.76 &
      75.82 &
      69.21 &
      72.36 &
      98.92 &
      94.21 &
      96.51 &
      95.87 &
      97.65 &
      \textbf{96.75} \\
    $\text{MetaPrompt}_{\text{ (TIP2024)}}$ &
      83.65 &
      75.48 &
      79.09 &
      77.52 &
      70.83 &
      74.02 &
      98.13 &
      94.58 &
      96.32 &
      95.53 &
      97.00 &
      96.26 \\
    $\text{TCP}_{\text{ (CVPR2024)}}$ &
      84.13 &
      75.36 &
      79.51 &
      77.27 &
      69.87 &
      73.38 &
      98.23 &
      \textbf{94.67} &
      96.42 &
      94.67 &
      97.20 &
      95.92 \\
    $\text{MMA}_{\text{ (CVPR2024)}}$ &
      83.20 &
      76.80 &
      79.87 &
      77.31 &
      71.00 &
      74.02 &
      98.40 &
      94.00 &
      96.15 &
      95.40 &
      \textbf{98.07} &
      96.72 \\ \midrule
    $\text{MMRL}_{\text{ (Ours)}}$ &
      \textbf{85.68} &
      \textbf{77.16} &
      \textbf{81.20} &
      \textbf{77.90} &
      \textbf{71.30} &
      \textbf{74.45} &
      \textbf{98.97} &
      94.50 &
      \textbf{96.68} &
      \textbf{95.90} &
      97.60 &
      96.74 \\ \midrule \midrule
    \multirow{2}{*}{Method} &
      \multicolumn{3}{c|}{StanfordCars} &
      \multicolumn{3}{c|}{Flowers102} &
      \multicolumn{3}{c|}{Food101} &
      \multicolumn{3}{c}{FGVCAircraft} \\
     &
      Base &
      Novel &
      HM &
      Base &
      Novel &
      HM &
      Base &
      Novel &
      HM &
      Base &
      Novel &
      HM \\ \midrule
    $\text{CLIP}_{\text{ (ICML2021)}}$ &
      63.37 &
      74.89 &
      68.65 &
      72.08 &
      77.80 &
      74.83 &
      90.10 &
      91.22 &
      90.66 &
      27.19 &
      36.29 &
      31.09 \\
    $\text{CoOp}_{\text{ (IJCV2022)}}$ &
      78.12 &
      60.40 &
      68.13 &
      97.60 &
      59.67 &
      74.06 &
      88.33 &
      82.26 &
      85.19 &
      40.44 &
      22.30 &
      28.75 \\
    $\text{CoOpOp}_{\text{ (CVPR2022)}}$ &
      70.49 &
      73.59 &
      72.01 &
      94.87 &
      71.75 &
      81.71 &
      90.70 &
      91.29 &
      90.99 &
      33.41 &
      23.71 &
      27.74 \\
    $\text{ProDA}_{\text{ (CVPR2022)}}$ &
      74.70 &
      71.20 &
      72.91 &
      97.70 &
      68.68 &
      80.66 &
      90.30 &
      88.57 &
      89.43 &
      36.90 &
      34.13 &
      35.46 \\
    $\text{KgCoOp}_{\text{ (CVPR2023)}}$ &
      71.76 &
      75.04 &
      73.36 &
      95.00 &
      74.73 &
      83.65 &
      90.50 &
      91.70 &
      91.09 &
      36.21 &
      33.55 &
      34.83 \\
    $\text{MaPLe}_{\text{ (CVPR2023)}}$ &
      72.94 &
      74.00 &
      73.47 &
      95.92 &
      72.46 &
      82.56 &
      90.71 &
      \textbf{92.05} &
      \textbf{91.38} &
      37.44 &
      35.61 &
      36.50 \\
    $\text{PromptSRC}_{\text{ (ICCV2023)}}$ &
      78.27 &
      74.97 &
      76.58 &
      98.07 &
      76.50 &
      85.95 &
      90.67 &
      91.53 &
      91.10 &
      42.73 &
      \textbf{37.87} &
      40.15 \\
    $\text{ProVP}_{\text{ (IJCV2024)}}$ &
      80.43 &
      67.96 &
      73.67 &
      98.42 &
      72.06 &
      83.20 &
      90.32 &
      90.91 &
      90.61 &
      \textbf{47.08} &
      29.87 &
      36.55 \\
    $\text{MetaPrompt}_{\text{ (TIP2024)}}$ &
      76.34 &
      75.01 &
      75.48 &
      97.66 &
      74.49 &
      84.52 &
      \textbf{90.74} &
      91.85 &
      91.29 &
      40.14 &
      36.51 &
      38.24 \\
    $\text{TCP}_{\text{ (CVPR2024)}}$ &
      80.80 &
      74.13 &
      77.32 &
      97.73 &
      75.57 &
      85.23 &
      90.57 &
      91.37 &
      90.97 &
      41.97 &
      34.43 &
      37.83 \\
    $\text{MMA}_{\text{ (CVPR2024)}}$ &
      78.50 &
      73.10 &
      75.70 &
      97.77 &
      75.93 &
      85.48 &
      90.13 &
      91.30 &
      90.71 &
      40.57 &
      36.33 &
      38.33 \\ \midrule
    $\text{MMRL}_{\text{ (Ours)}}$ &
      \textbf{81.30} &
      \textbf{75.07} &
      \textbf{78.06} &
      \textbf{98.97} &
      \textbf{77.27} &
      \textbf{86.78} &
      90.57 &
      91.50 &
      91.03 &
      46.30 &
      37.03 &
      \textbf{41.15} \\ \midrule \midrule
    \multirow{2}{*}{Method} &
      \multicolumn{3}{c|}{SUN397} &
      \multicolumn{3}{c|}{DTD} &
      \multicolumn{3}{c|}{EuroSAT} &
      \multicolumn{3}{c}{UCF101} \\
     &
      Base &
      Novel &
      HM &
      Base &
      Novel &
      HM &
      Base &
      Novel &
      HM &
      Base &
      Novel &
      HM \\ \midrule
    $\text{CLIP}_{\text{ (ICML2021)}}$ &
      69.36 &
      75.35 &
      72.23 &
      53.24 &
      59.90 &
      56.37 &
      56.48 &
      64.05 &
      60.03 &
      70.53 &
      77.50 &
      73.85 \\
    $\text{CoOp}_{\text{ (IJCV2022)}}$ &
      80.60 &
      65.89 &
      72.51 &
      79.44 &
      41.18 &
      54.24 &
      92.19 &
      54.74 &
      68.69 &
      84.69 &
      56.05 &
      67.46 \\
    $\text{CoOpOp}_{\text{ (CVPR2022)}}$ &
      79.74 &
      76.86 &
      78.27 &
      77.01 &
      56.00 &
      64.85 &
      87.49 &
      60.04 &
      71.21 &
      82.33 &
      73.45 &
      77.64 \\
    $\text{ProDA}_{\text{ (CVPR2022)}}$ &
      78.67 &
      76.93 &
      77.79 &
      80.67 &
      56.48 &
      66.44 &
      83.90 &
      66.00 &
      73.88 &
      85.23 &
      71.97 &
      78.04 \\
    $\text{KgCoOp}_{\text{ (CVPR2023)}}$ &
      80.29 &
      76.53 &
      78.36 &
      77.55 &
      54.99 &
      64.35 &
      85.64 &
      64.34 &
      73.48 &
      82.89 &
      76.67 &
      79.65 \\
    $\text{MaPLe}_{\text{ (CVPR2023)}}$ &
      80.82 &
      78.70 &
      79.75 &
      80.36 &
      59.18 &
      68.16 &
      94.07 &
      73.23 &
      82.35 &
      83.00 &
      78.66 &
      80.77 \\
    $\text{PromptSRC}_{\text{ (ICCV2023)}}$ &
      82.67 &
      78.47 &
      80.52 &
      83.37 &
      62.97 &
      71.75 &
      92.90 &
      73.90 &
      82.32 &
      87.10 &
      78.80 &
      82.74 \\
    $\text{ProVP}_{\text{ (IJCV2024)}}$ &
      80.67 &
      76.11 &
      78.32 &
      83.95 &
      59.06 &
      69.34 &
      \textbf{97.12} &
      72.91 &
      83.29 &
      \textbf{88.56} &
      75.55 &
      81.54 \\
    $\text{MetaPrompt}_{\text{ (TIP2024)}}$ &
      82.26 &
      79.04 &
      80.62 &
      83.10 &
      58.05 &
      68.35 &
      93.53 &
      75.21 &
      83.38 &
      85.33 &
      77.72 &
      81.35 \\
    $\text{TCP}_{\text{ (CVPR2024)}}$ &
      82.63 &
      78.20 &
      80.35 &
      82.77 &
      58.07 &
      68.25 &
      91.63 &
      74.73 &
      82.32 &
      87.13 &
      \textbf{80.77} &
      83.83 \\
    $\text{MMA}_{\text{ (CVPR2024)}}$ &
      82.27 &
      78.57 &
      80.38 &
      83.20 &
      \textbf{65.63} &
      73.38 &
      85.46 &
      \textbf{82.34} &
      83.87 &
      86.23 &
      80.03 &
      82.20 \\ \midrule
    $\text{MMRL}_{\text{ (Ours)}}$ &
      \textbf{83.20} &
      \textbf{79.30} &
      \textbf{81.20} &
      \textbf{85.67} &
      65.00 &
      \textbf{73.82} &
      95.60 &
      80.17 &
      \textbf{87.21} &
      88.10 &
      80.07 &
      \textbf{83.89} \\ \bottomrule
    \end{tabular}
            }
}
\vspace{-0.3cm}
\end{table*}

\section{Experiments}
\label{sec:experiments}
\textbf{Details on implementation, datasets, and computational cost are provided in the Supplementary Materials}.

\subsection{Tasks and Datasets}
We conduct four core evaluations to comprehensively assess MMRL's performance: base-to-novel generalization, cross-dataset evaluation, domain generalization, and few-shot learning. Except for few-shot learning, all experiments utilize a 16-shot setting, \ie., only 16 training examples per category.

\noindent \textbf{Base-to-Novel Generalization:} 
In this evaluation, dataset classes are equally divided into base and novel classes. The model is trained exclusively on base classes and tested on both base and novel classes, allowing us to examine its transfer learning effectiveness on base classes as well as its ability to retain the inherent generalization or zero-shot capabilities of pre-trained VLMs for novel classes. We conduct this evaluation across 11 diverse image classification datasets: ImageNet \cite{imagenet}, Caltech101 \cite{caltech101}, OxfordPets \cite{oxford_pets}, StanfordCars \cite{stanford_cars}, Flowers102 \cite{flowers102}, Food101 \cite{food101}, FGVCAircraft \cite{fgvc_aircraft}, SUN397 \cite{sun397}, UCF101 \cite{ucf101}, DTD \cite{dtd}, and EuroSAT \cite{eurosat}.

\noindent \textbf{Cross-Dataset Evaluation:} This evaluation measures the model’s transferability to new, unseen datasets. Following CoCoOp \cite{cocoop}, we train the model on all 1000 ImageNet classes in a few-shot setting and then directly apply it, without further fine-tuning, to other datasets to assess its cross-dataset generalization. We employ the same datasets as in the base-to-novel generalization task.

\noindent \textbf{Domain Generalization:} In this setting, we assess the resilience of the ImageNet-trained model to domain shifts and its generalization to out-of-distribution data. Specifically, we use ImageNet as the training dataset and evaluate on four variants—ImageNetV2 \cite{imagenetv2}, ImageNet-Sketch \cite{imagenet_sketch}, ImageNet-A \cite{imagenet_a}, and ImageNet-R \cite{imagenet_r}—each introducing different types of domain variation.

\noindent \textbf{Few-Shot Learning:} This evaluation examines the model's transfer learning capability in limited-data scenarios, independent of its generalization performance. The model is trained on subsets of the training data with 1, 2, 4, 8, and 16 examples (shots) per class and subsequently evaluated on the full test sets.

\begin{table*}[t]
\centering
\setlength{\abovecaptionskip}{0.15cm}  
\caption{Comparison of MMRL with previous state-of-the-art methods on cross-dataset evaluation across 10 datasets.}
\label{cross_dataset}
\resizebox{0.9\textwidth}{!}{
    \footnotesize
    \begin{tabular}{@{}rc|ccccccccccc@{}}
    \toprule
     &
      Source &
      \multicolumn{11}{c}{Target} \\ \cmidrule(l){2-13} 
     &
      \rotatebox{60}{ImageNet} &
      \rotatebox{60}{Average} &
      \rotatebox{60}{Caltech101} &
      \rotatebox{60}{OxfordPets} &
      \rotatebox{60}{StanfordCars} &
      \rotatebox{60}{Flowers101} &
      \rotatebox{60}{Food101} &
      \rotatebox{60}{FGVCAircraft} &
      \rotatebox{60}{SUN397} &
      \rotatebox{60}{DTD} &
      \rotatebox{60}{EuroSAT} &
      \rotatebox{60}{UCF101} \\ \cmidrule(l){2-13} 
    $\text{CoOp}_{\text{ (IJCV2022)}}$ &
      71.51 &
      63.88 &
      93.70 &
      89.14 &
      64.51 &
      68.71 &
      85.30 &
      18.47 &
      64.15 &
      41.92 &
      46.39 &
      66.55 \\
    $\text{CoOpOp}_{\text{ (CVPR2022)}}$ &
      71.02 &
      65.74 &
      94.43 &
      90.14 &
      65.32 &
      71.88 &
      86.06 &
      22.94 &
      67.36 &
      45.73 &
      45.37 &
      68.21 \\
    $\text{MaPLe}_{\text{ (CVPR2023)}}$ &
      70.72 &
      66.30 &
      93.53 &
      90.49 &
      65.57 &
      72.23 &
      86.20 &
      24.74 &
      67.01 &
      46.49 &
      48.06 &
      68.69 \\
    $\text{PromptSRC}_{\text{ (ICCV2023)}}$ &
      71.27 &
      65.81 &
      93.60 &
      90.25 &
      65.70 &
      70.25 &
      86.15 &
      23.90 &
      67.10 &
      \textbf{46.87} &
      45.50 &
      \textbf{68.75} \\
    $\text{TCP}_{\text{ (CVPR2024)}}$ &
      71.40 &
      66.29 &
      93.97 &
      91.25 &
      64.69 &
      71.21 &
      \textbf{86.69} &
      23.45 &
      67.15 &
      44.35 &
      51.45 &
      68.73 \\
    $\text{MMA}_{\text{ (CVPR2024)}}$ &
      71.00 &
      66.61 &
      93.80 &
      90.30 &
      \textbf{66.13} &
      72.07 &
      86.12 &
      25.33 &
      \textbf{68.17} &
      46.57 &
      49.24 &
      68.32 \\ \midrule
    $\text{MMRL}_{\text{ (Ours)}}$ &
      \textbf{72.03} &
      \textbf{67.25} &
      \textbf{94.67} &
      \textbf{91.43} &
      66.10 &
      \textbf{72.77} &
      86.40 &
      \textbf{26.30} &
      67.57 &
      45.90 &
      \textbf{53.10} &
      68.27 \\ \bottomrule
    \end{tabular}
    }
\vspace{-0.3cm}
\end{table*}

\begin{figure*}[t]
\centering
\setlength{\belowcaptionskip}{-0.39cm}  
  \includegraphics[width=0.9\linewidth]{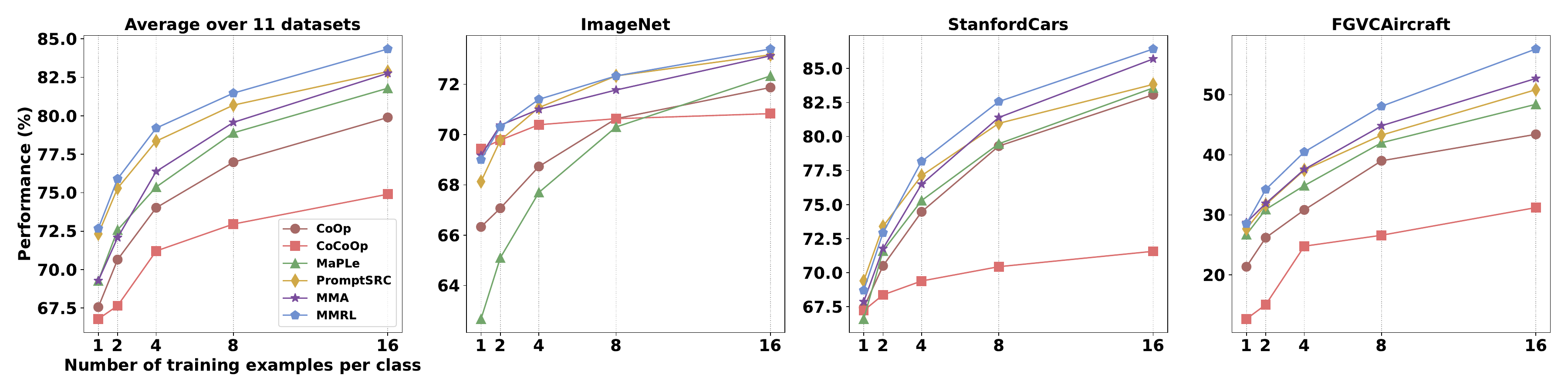}
  \vspace{-0.3cm}
  \caption{Comparison of MMRL with previous state-of-the-art methods on few-shot learning across 11 datasets. Detailed results on all 11 datasets are provided in the \textbf{Supplementary Material}.}
  \label{few_shot_figure}
\end{figure*}

\subsection{Base-to-Novel Generalization}

In this experiment, we compare MMRL with several models, including the zero-shot baseline CLIP and leading prompt learning approaches: CoOp \cite{coop}, CoCoOp \cite{cocoop}, ProDA \cite{proda}, KgCoOp \cite{kgcoop}, MaPLe \cite{maple}, PromptSRC \cite{promptsrc}, ProVP \cite{provp}, MetaPrompt \cite{metaprompt}, TCP \cite{tcp} and the multimodal adapter-style model MMA \cite{mma}. 

\cref{base_to_novel} provides detailed results for \textbf{Base} and \textbf{Novel} classes across 11 datasets, along with the balanced harmonic mean (\textbf{HM}) of their accuracies. Key findings include:

\noindent \textbf{New SOTA Performance:} Based on the average results across 11 datasets, MMRL achieves gains of 2.48\%, 0.36\%, and 1.33\% in Base, Novel, and HM metrics, respectively, surpassing the previous best-performing model, MMA, and establishing a new state-of-the-art.

\noindent \textbf{Strong Generalizability with Enhanced Transfer Learning:} Notably, MMRL enhances generalizability while significantly boosting base accuracy, effectively improving transfer learning capabilities across downstream datasets such as ImageNet, StanfordCars, and SUN397. Although MMRL may not consistently achieve the highest novel accuracy on some datasets (e.g., UCF101, EuroSAT, DTD, and FGVCAircraft), it substantially outperforms other methods in the base category. For instance, on FGVCAircraft, MMRL’s novel accuracy trails PromptSRC by 0.84\%, yet it achieves a significant 3.57\% gain in base accuracy. Similarly, on EuroSAT, MMRL underperforms MMA by 2.17\% in the novel category but outperforms it by 10.14\% in the base category!

\begin{table}[t]
\centering
\setlength{\abovecaptionskip}{0.15cm}  
\caption{Comparison of MMRL with previous state-of-the-art methods on domain generalization across 4 datasets.}
\label{domain_generalization}
\resizebox{0.45\textwidth}{!}{
    \footnotesize
    \begin{tabular}{@{}rc|cccc@{}}
    \toprule
                                         & Source   & \multicolumn{4}{c}{Target}    \\ \cmidrule(l){2-6} 
                                         & ImageNet & -V2   & -S    & -A    & -R    \\ \cmidrule(l){2-6} 
    $\text{CLIP}_{\text{ (ICML2021)}}$   & 66.73    & 60.83 & 46.15 & 47.77 & 73.96 \\
    $\text{CoOp}_{\text{ (IJCV2022)}}$   & 71.51    & 64.20 & 47.99 & 49.71 & 75.21 \\
    $\text{CoOpOp}_{\text{ (CVPR2022)}}$ & 71.02    & 64.07 & 48.75 & 50.63 & 76.18 \\
    $\text{MaPLe}_{\text{ (CVPR2023)}}$  & 70.72    & 64.07 & 49.15 & 50.90 & 76.98 \\
    $\text{PromptSRC}_{\text{ (ICCV2023)}}$ & 71.27          & 64.35          & \textbf{49.55} & 50.90          & \textbf{77.80} \\
    $\text{MMA}_{\text{ (CVPR2024)}}$    & 71.00    & 64.33 & 49.13 & 51.12 & 77.32 \\ \midrule
    $\text{MMRL}_{\text{ (Ours)}}$          & \textbf{72.03} & \textbf{64.47} & 49.17          & \textbf{51.20} & 77.53 \\ \bottomrule
    \end{tabular}
    }
\vspace{-0.4cm}
\end{table}

\subsection{Cross-Dataset Evaluation}
As illustrated in \cref{cross_dataset}, MMRL achieves a 1.03\% accuracy improvement over the previous state-of-the-art method, MMA, on ImageNet. Beyond this, MMRL consistently exhibits superior performance across various target datasets, achieving the highest average accuracy, which underscores its strong cross-dataset generalization capability.

\subsection{Domain Generalization}
As summarized in \cref{domain_generalization}, MMRL attains top performance on 2 out of the 4 domain-shifted datasets, showcasing its robust generalization capability across diverse domains.

\subsection{Few-Shot Learning}
As shown in \cref{few_shot_figure}, MMRL achieves the best average performance across 11 datasets under all shot settings, with performance margins increasing as the shot number rises. This trend confirms MMRL’s strong transfer learning capability, even in data-scarce scenarios.

\begin{figure}[tb]
\centering
\setlength{\abovecaptionskip}{0.15cm}   
  \includegraphics[width=1.0\linewidth]{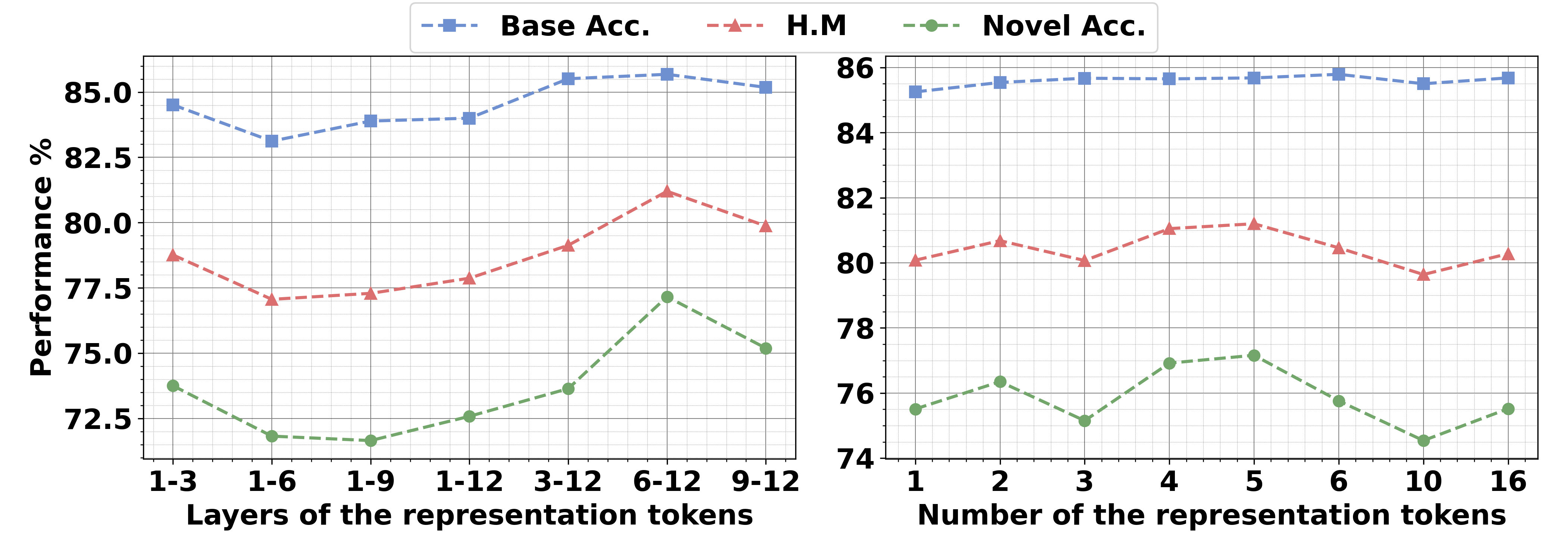}
  \caption{Ablation on layers (left) and $K$ (right).}
  \label{ablation_layers_numbers}
\vspace{-0.5cm}
\end{figure}

\begin{table}[pb]
\centering
\vspace{-0.4cm}
\setlength{\abovecaptionskip}{0.15cm}  
\caption{Ablation on variants (left) and $d_r$ (right).}
\label{ablation_variants_dr}
    \begin{subtable}{0.495\linewidth}
    \centering
    \resizebox{0.95\textwidth}{!}{
        \footnotesize
        \begin{tabular}{@{}c|ccc@{}}
        \toprule
        Variants              & Base           & Novel          & HM             \\ \midrule
        w/o L                 & 85.05          & 75.65          & 80.08          \\
        w/o V                 & 82.83          & 75.03          & 78.74          \\
        w/o $\text{DS}_1$     & 83.59          & 77.16          & 80.25          \\
        w/o $\text{DS}_2$     & 85.68          & 73.80          & 79.30          \\
        w/o RS                & \textbf{85.79} & 75.55          & 80.34          \\
        $\text{MMRL}^\dagger$ & 85.60          & 76.02          & 80.55          \\
        \rowcolor[HTML]{EFEFEF} 
        MMRL                  & 85.68          & \textbf{77.16} & \textbf{81.20} \\ \bottomrule
        \end{tabular}
        }
    \end{subtable}
    \begin{subtable}{0.495\linewidth}
    \centering
    \resizebox{0.965\textwidth}{!}{
        \footnotesize
        \begin{tabular}{@{}c|ccc@{}}
        \toprule
        $d_r$ & Base           & Novel          & HM             \\ \midrule
        32    & 85.27          & 76.85          & 80.84          \\
        128   & 85.42          & 76.74          & 80.85          \\
        256   & 85.63          & 76.84          & 81.00          \\
        \rowcolor[HTML]{EFEFEF} 
        512   & \textbf{85.68} & \textbf{77.16} & \textbf{81.20} \\
        1024  & 85.57          & 76.97          & 81.04          \\
        2048  & 85.53          & 76.91          & 81.00          \\ \bottomrule
        \end{tabular}
        }
    \end{subtable}

\end{table}

\subsection{Ablation Analysis}
All ablation experiments are conducted on base-to-novel generalization across 11 datasets, with results averaged, except for the analysis of $\lambda$ on ImageNet; please refer to the \textbf{Supplementary Material} for the complete $\lambda$ analysis across all datasets.

\noindent \textbf{Variants of MMRL:} The performance of MMRL variants, as shown in \cref{ablation_variants_dr} (left), highlights the contributions of different components. In the variants `w/o L' and `w/o V', where only a single modality of representation tokens is utilized, we observe a performance decline, especially in `w/o V', emphasizing the significance of multimodal interaction and representation features in MMRL. The `w/o $\text{DS}_1$' variant, which omits the Decoupling Strategy and relies only on class features, degrades significantly on base classes, underscoring the role of representation features in capturing downstream knowledge. The `w/o $\text{DS}_2$' variant, using both features for novel class evaluation, also drops notably, suggesting that the representation tokens primarily capture base class features, making transfer to new tasks challenging. The `w/o RS' variant, which excludes the Representation Space, independently initializes textual and visual tokens without multimodal learning. This unimodal approach, while improving base class performance, severely limits generalization to novel classes, indicating the necessity of multimodal learning for effective generalization. Finally, $\text{MMRL}^\dagger$ adopts a biased multimodal learning scheme similar to MaPLe \cite{maple}, where text-side tokens are initialized randomly and mapped to the visual side. Results indicate that this biased approach underperforms compared to MMRL's balanced multimodal learning framework.

\noindent \textbf{Dimension of Representation Space, $d_r$:} As shown in \cref{ablation_variants_dr} (right), adjusting $d_r$ reveals that performance initially increases, followed by a decline as $d_r$ continues to grow. This decline likely stems from overfitting caused by an overly complex representation space.

\noindent \textbf{Layer for Representation Token Insertion:} As depicted in \cref{ablation_layers_numbers} (left), model performance declines when representation tokens are introduced at lower encoder layers. This trend aligns with MMRL's design, as higher layers capture dataset-specific, discriminative features, while lower layers retain generalizable features. Furthermore, MMRL performs poorly on base classes when representation tokens are inserted into lower layers, suggesting that lower-layer features are less adaptable. Performance improves with insertion at higher layers but declines when placed too high, likely due to the reduced number of learnable parameters and limited capacity to influence CLIP's critical parameters.

\begin{table}[t]
\centering
\setlength{\abovecaptionskip}{0.15cm}  
\caption{Ablation on $\alpha$ (left) across 11 datasets and $\lambda$ (right) on ImageNet.}
\label{ablation_alpha_lambda}
    \begin{subtable}{0.46\linewidth}
    \centering
    \resizebox{0.9\textwidth}{!}{
        \footnotesize
        \begin{tabular}{@{}c|ccc@{}}
        \toprule
        $\alpha$ & Base           & Novel          & HM             \\ \midrule
        0.0      & 82.96          & 72.34          & 77.29          \\
        0.3      & 84.57          & 75.45          & 79.75          \\
        0.5      & 85.42          & 76.11          & 80.50          \\
        \rowcolor[HTML]{EFEFEF} 
        0.7      & \textbf{85.68} & \textbf{77.16} & \textbf{81.20} \\
        1.0      & 83.79          & 75.49          & 79.42          \\ \bottomrule
        \end{tabular}
        }
    \end{subtable}
    \begin{subtable}{0.46\linewidth}
    \centering
    \resizebox{0.9\textwidth}{!}{
        \footnotesize
        \begin{tabular}{@{}c|ccc@{}}
        \toprule
        $\lambda$ & Base           & Novel          & HM             \\ \midrule
        0.0       & 77.73          & 70.63          & 73.96          \\
        0.2       & 77.83          & 71.23          & 74.38          \\
        \rowcolor[HTML]{EFEFEF} 
        0.5       & \textbf{77.90} & \textbf{71.30} & \textbf{74.45} \\
        2.0       & \textbf{77.90} & 70.93          & 74.25          \\
        4.0       & 77.67          & 70.73          & 74.04          \\ \bottomrule
        \end{tabular}
        }
    \end{subtable}
\vspace{-0.4cm}
\end{table}

\noindent \textbf{Number of Representation Tokens, $K$:} In \cref{ablation_layers_numbers} (right), increasing $K$ slightly enhances base class accuracy due to additional learnable parameters. For novel classes, accuracy initially improves with $K$ but eventually declines, indicating that an excessive number of tokens may lead to overfitting, reducing generalization capacity.

\noindent \textbf{Balance Weight, $\alpha$:} The parameter $\alpha$ modulates reliance on representation token features versus class token features. Lower $\alpha$ values increase dependence on representation features, heightening overfitting risk due to the learnable projection layer, while higher values shift dependence to class token features, diminishing transferability. As shown in \cref{ablation_alpha_lambda} (left), the optimal $\alpha$ is 0.7.

\noindent \textbf{Penalty Coefficient, $\lambda$:}  The penalty coefficient $\lambda$ regulates the regularization strength by aligning class token features with frozen CLIP’s features. Higher $\lambda$ generally enhances generalization but may restrict transfer flexibility. In \cref{ablation_alpha_lambda} (right), the optimal $\lambda$ for ImageNet is 0.5.

\section{Conclusion}
In this work, we introduce the MMRL framework to enhance the generalization of VLMs when adapting to diverse downstream datasets. MMRL establishes a shared, unbiased representation space that bridges image and text modalities, promoting balanced multimodal learning while preserving the pre-trained knowledge encapsulated in class tokens. By strategically decoupling representation tokens from class tokens during inference, MMRL effectively mitigates overfitting risks and reinforces adaptability. Extensive evaluations confirm MMRL’s capacity for an optimal balance between task-specific adaptation and generalization, setting a new benchmark for efficient transfer learning. 
\section*{Acknowledgement}
\label{sec:acknowledgement}
This work was supported in part by National Natural Science Foundation of China under Grant 62176062.

{
    \small
    \bibliographystyle{ieeenat_fullname}
    \bibliography{main}
}
\clearpage
\appendix
\setcounter{page}{1}
\maketitlesupplementary

\section{Implementation Details}
We follow prior studies\cite{coop, cocoop, prograd, kgcoop, maple, tcp, mma} and adopt a 16-shot learning setting across all experiments, except for the few-shot learning tasks. The ViT-B/16\cite{vit} variant of the CLIP model serves as the visual backbone for all experimental setups. Hand-crafted text prompts from prior methods\cite{clip, coop, tip-adapter} are utilized and described in detail in \cref{datasets}. Optimization is performed using the AdamW optimizer with an initial learning rate of 0.001. All our models are trained with mix-precision for speeding up. For the larger ImageNet dataset, we employ a batch size of 32, while a batch size of 4 is used for all other datasets. Training on ImageNet for the base-to-novel generalization task spans 5 epochs, whereas training on the remaining datasets is conducted over 10 epochs. For cross-dataset evaluation and domain generalization tasks, we perform training for a single epoch on ImageNet. In the few-shot learning tasks, training is carried out for 5 epochs on ImageNet and 50 epochs for other datasets. The average accuracy is reported over three independent runs, with all experiments executed on a single NVIDIA RTX 4090 GPU.

Representation tokens are initialized from a zero-mean Gaussian distribution with a standard deviation of 0.02. We set $J = 6$, integrating the representation tokens beginning at the 6-th transformer layer. The dimension of the representation space, $d_r$, is set to 2048 for EuroSAT and 512 for all other datasets. Note that since the $d_r$ setting for EuroSAT differs from other datasets, in the $d_r$ ablation experiments we fix $d_r$ for EuroSAT to 2048 while adjusting $d_r$ on the other datasets. The number of representation tokens, $K$, is configured to 5. The parameter $\alpha$ is fixed at 0.7, and the details regarding the configuration of $\lambda$ are provided in \cref{ablation_lambda}.

\section{Dataset Details}
Details of 14 datasets are shown in \cref{datasets}.

\section{Computational Cost}
Table \ref{computational_cost} summarizes the learnable parameters, training time per image, total training duration, inference speed (measured in frames per second, FPS, with a batch size of 100), and the final HM metric for each approach. Our proposed model, MMRL, demonstrates a compelling balance of computational efficiency and performance. The key observations are as follows:
\begin{itemize}
    \item Models incorporating multimodal interaction mechanisms (e.g., MaPLe, MMA, and MMRL) generally involve a higher parameter count compared to models without such mechanisms.
    \item Both MMRL and the prior MMA approach exhibit significantly faster training speed, thereby reducing overall computational costs. While MaPLe and PromptSRC achieve higher inference speeds, their training durations are relatively longer. Notably, MMRL offers faster inference compared to MMA and MetaPrompt.
    \item To assess the performance of MMRL under constrained computational resources, we reduced the dimensionality of the representation space from 512 to 32. In this configuration, MMRL achieves a parameter count comparable to that of MMA, while still significantly outperforming the previous state-of-the-art model.
\end{itemize}

\begin{table}[h]
\centering
\renewcommand\arraystretch{1.25}
\caption{All methods were trained on a single NVIDIA RTX 4090 GPU using the ImageNet dataset. Each model was implemented with publicly available code and default configurations as described in their respective papers \cite{maple, promptsrc, provp, metaprompt, tcp, mma}. `V-L' denotes vision-language interaction, indicating that efficient fine-tuning incorporates interactions between visual and textual modalities before prediction. `V, L' signifies separate fine-tuning of each modality without inter-modal interaction before prediction, while `L' refers to fine-tuning limited to the textual modality alone. `Train time' is reported as both time per image and the total duration for training the full dataset(16-shots), while `FPS (100 BS)' indicates frames per second with a batch size of 100 during inference.}
\label{computational_cost}
\resizebox{0.475\textwidth}{!}{
    \begin{tabular}{@{}l|ccccc|c@{}}
    \toprule
    \multirow{2}{*}{Method} & \multirow{2}{*}{Modality} & Params & Train time & Train time & FPS & \multirow{2}{*}{HM} \\
               &     & (learnable) & (ms/image) & (minute/all) & (100 BS) &       \\ \midrule
    MaPLe      & V-L & 3.555M      & 39.5       & 26.4         & 1757.6   & 78.55 \\
    PromptSRC  & V,L & 0.046M      & 40.0       & 106.8        & 1764.2   & 79.97 \\
    ProVP      & V   & 0.147M      & 4.4        & 107.2        & 928.9    & 78.76 \\
    MetaPrompt & V,L & 0.031M      & 30.7       & 32.8         & 659.8    & 79.09 \\
    TCP        & L   & 0.332M      & 5.3        & 17.7         & 950.6    & 79.51 \\
    MMA        & V-L & 0.675M      & 2.2        & 1.5          & 688.5    & 79.87 \\ \midrule
    MMRL       & V-L & 4.992M      & 5.3        & 3.6          & 762.4    & 81.20 \\
    MMRL*      & V-L & 0.689M      & 5.3        & 3.6          & 767.8    & 80.84 \\ \bottomrule
    \end{tabular}
}
\end{table}

\section{Ablation Analysis on $\lambda$}
As shown in \cref{ablation_lambda}, increasing the value of $\lambda$ generally improves performance, with the optimal or near-optimal results typically observed when $\lambda$ is set between 4 and 6 across most datasets. Notably, as $\lambda$ continues to increase, its impact on model performance within the same dataset diminishes, indicating reduced sensitivity to variations in $\lambda$. This trend suggests that the model becomes more robust and less reliant on precise tuning of $\lambda$ at higher values.

\begin{table*}[h]
\centering
\caption{Summary of the 14 datasets.}
\label{datasets}
\renewcommand\arraystretch{1.2}
\resizebox{1.0\textwidth}{!}{
    \begin{tabular}{@{}l|llllll@{}}
    \toprule
    Dataset      & Classes & Train  & Val    & Test   & Description                         & Prompt                                \\ \midrule
    ImageNet     & 1000    & 1.28M  & $\sim$ & 50000  & Recognition of generic objects      & ``a photo of a [CLASS].”              \\
    Caltech101   & 100     & 4128   & 1649   & 2465   & Recognition of generic objects      & ``a photo of a [CLASS].”              \\
    OxfordPets      & 37    & 2944   & 736    & 3669   & Fine-grained classification of pets                    & ``a photo of a [CLASS], a type of pet.”      \\
    StanfordCars & 196     & 6509   & 1635   & 8041   & Fine-grained classification of cars & ``a photo of a [CLASS].”              \\
    Flowers102      & 102   & 4093   & 1633   & 2463   & Fine-grained classification of flowers                 & ``a photo of a [CLASS], a type of flower.”   \\
    Food101         & 101   & 50500  & 20200  & 30300  & Fine-grained classification of foods                   & ``a photo of [CLASS], a type of food.”       \\
    FGVCAircraft    & 100   & 3334   & 3333   & 3333   & Fine-grained classification of aircrafts               & ``a photo of a [CLASS], a type of aircraft.” \\
    SUN397       & 397     & 15880  & 3970   & 19850  & Scene classification                & ``a photo of a [CLASS].”              \\
    DTD          & 47      & 2820   & 1128   & 1692   & Texture classification              & ``[CLASS] texture.”                   \\
    EuroSAT         & 10    & 13500  & 5400   & 8100   & Land use \& cover classification with satellite images & ``a centered satellite photo of [CLASS].”    \\
    UCF101       & 101     & 7639   & 1898   & 3783   & Action recognition                  & ``a photo of a person doing [CLASS].” \\ \midrule
    ImageNetV2   & 1,000   & $\sim$ & $\sim$ & 10,000 & New test data for ImageNet          & ``a photo of a [CLASS].”              \\
    ImageNet-Sketch & 1,000 & $\sim$ & $\sim$ & 50,889 & Sketch-style images of ImageNet classes                & ``a photo of a [CLASS].”                     \\
    ImageNet-A      & 200   & $\sim$ & $\sim$ & 7,500  & Natural adversarial examples of 200 ImageNet classes   & ``a photo of a [CLASS].”                     \\
    ImageNet-R   & 200     & $\sim$ & $\sim$ & 30,000 & Renditions of 200 ImageNet classes  & ``a photo of a [CLASS].”              \\ \bottomrule
    \end{tabular}
    }
\end{table*}

\begin{table*}[h]
\centering
\caption{Ablation on $\lambda$ across 11 datasets, with results evaluated using the harmonic mean (HM) metric.}
\label{ablation_lambda}
\renewcommand\arraystretch{1.2}
\resizebox{1.0\textwidth}{!}{
    \begin{tabular}{@{}c|ccccccccccc@{}}
    \toprule
    $\alpha$ & ImageNet       & Caltech101     & OxfordPets & StanfordCars & Flowers102 & Food101 & FGVCAircraft   & SUN397         & DTD            & EuroSAT & UCF101 \\ \midrule
    0.0  & 74.01 & 95.97 & 96.35          & 76.00          & 84.42          & 90.10          & 38.52 & 79.67 & 68.21 & 82.65          & 81.63          \\
    0.01 & 74.07 & 96.12 & 96.39          & 75.95          & 84.82          & 90.23          & 37.87 & 79.85 & 67.73 & \textbf{87.21} & 82.11          \\
    0.1  & 74.23 & 96.25 & 96.49          & 76.32          & 84.81          & 90.53          & 38.66 & 80.23 & 69.79 & 83.21          & 82.91          \\
    0.2  & 74.38 & 96.40 & \textbf{96.74} & 76.67          & 85.31          & 90.61          & 39.27 & 80.25 & 70.58 & 82.68          & 82.70          \\
    0.5      & \textbf{74.45} & \textbf{96.68} & 96.54      & 77.09        & 85.74      & 90.86   & 40.37          & 80.61          & 72.67          & 82.87   & 83.05  \\
    3.0  & 74.09 & 96.59 & 96.51          & 77.72          & 86.65          & 90.98          & 40.48 & 81.10 & 73.54 & 77.95          & \textbf{83.89} \\
    4.0  & 74.04 & 96.62 & 96.55          & 77.73          & \textbf{86.78} & 90.98          & 40.66 & 81.14 & 73.75 & 77.27          & 83.45          \\
    5.0  & 73.93 & 96.62 & 96.60          & 77.86          & 86.42          & \textbf{91.03} & 40.42 & 81.07 & 73.69 & 78.05          & 83.84          \\
    6.0      & 73.83          & 96.61          & 96.66      & 78.05        & 86.48      & 91.00   & \textbf{41.15} & \textbf{81.20} & \textbf{73.82} & 75.23   & 83.68  \\
    7.0  & 73.78 & 96.62 & 96.58          & \textbf{78.06} & 86.53          & 90.95          & 40.88 & 81.10 & 73.65 & 75.85          & 83.55          \\
    10.0 & 73.68 & 96.64 & 96.56          & 77.86          & 86.46          & 91.00          & 41.01 & 80.93 & 73.68 & 77.61          & 83.38          \\ \bottomrule
    \end{tabular}
}
\end{table*}

\begin{table}[t]
\small
\centering
\caption{Ablation on different regularization strategies.}
\label{ablation_regularization}
\begin{tabular}{@{}c|ccc@{}}
\toprule
Regularization & Base           & Novel          & HM             \\ \midrule
\rowcolor[HTML]{EFEFEF} 
Cosine         & \textbf{85.68} & \textbf{77.16} & \textbf{81.20} \\
L1             & 85.46          & 76.03          & 80.47          \\
MSE             & 85.13          & 74.62          & 79.53          \\ \bottomrule
\end{tabular}
\end{table}

\section{Ablation Analysis on Regularization Strategies}
We investigate the impact of various regularization strategies aimed at maximizing the similarity between class token features and frozen CLIP features to retain pre-trained knowledge. The results, summarized in \cref{ablation_regularization}, indicate that cosine regularization achieves the best performance. In contrast, both L1 and MSE losses lead to performance degradation, with MSE causing a significant decline. This result can be attributed to the more relaxed and flexible constraints of cosine regularization, enabling the class token to preserve generalizability while effectively capturing task-specific knowledge.

\section{Few-Shot Learning}
\cref{few_shot1,few_shot2} provide detailed comparisons of MMRL and prior state-of-the-art methods on few-shot learning across 11 datasets. MMRL achieves the highest average performance across all shots. Note that the MMA results are reproduced from the open-source code, as the original paper does not report results for this experiment.

\begin{table*}[t]
\small
\centering
\caption{Comparison of MMRL with previous state-of-the-art methods on few-shot learning across 11 datasets.}
\label{few_shot1}
\setlength{\tabcolsep}{15pt}{
\resizebox{0.9\textwidth}{!}{
    \begin{tabular}{@{}ll|ccccc}
    \toprule
    \textbf{Dataset} &
      \textbf{Method} &
      \textbf{1 shot} &
      \textbf{2 shots} &
      \textbf{4 shots} &
      \textbf{8 shots} &
      \textbf{16 shots} \\ \midrule
     &
      Linear probe CLIP &
      45.83 &
      57.98 &
      68.01 &
      74.47 &
      78.79 \\
     &
      CoOp &
      67.56 &
      70.65 &
      74.02 &
      76.98 &
      79.89 \\
     &
      CoCoOp &
      66.79 &
      67.65 &
      71.21 &
      72.96 &
      74.90 \\
     &
      MaPLe &
      69.27 &
      72.58 &
      75.37 &
      78.89 &
      81.79 \\
     &
      PromptSRC &
      72.32 &
      75.29 &
      78.35 &
      80.69 &
      82.87 \\
     &
      MMA &
      69.28 &
      72.08 &
      76.38 &
      79.57 &
      82.76 \\
    \multirow{-7}{*}{Average} &
      \cellcolor[HTML]{E8E8E8}$\text{MMRL}_{\text{ (Ours)}}$ &
      \cellcolor[HTML]{E8E8E8}\textbf{72.67} &
      \cellcolor[HTML]{E8E8E8}\textbf{75.90} &
      \cellcolor[HTML]{E8E8E8}\textbf{79.20} &
      \cellcolor[HTML]{E8E8E8}\textbf{81.47} &
      \cellcolor[HTML]{E8E8E8}\textbf{84.34} \\ \midrule
     &
      Linear probe CLIP &
      32.13 &
      44.88 &
      54.85 &
      62.23 &
      67.31 \\
     &
      CoOp &
      66.33 &
      67.07 &
      68.73 &
      70.63 &
      71.87 \\
     &
      CoCoOp &
      69.43 &
      69.78 &
      70.39 &
      70.63 &
      70.83 \\
     &
      MaPLe &
      62.67 &
      65.10 &
      67.70 &
      70.30 &
      72.33 \\
     &
      PromptSRC &
      68.13 &
      69.77 &
      71.07 &
      \textbf{72.33} &
      73.17 \\
     &
      MMA &
      \textbf{69.17} &
      \textbf{70.37} &
      71.00 &
      71.77 &
      73.13 \\
    \multirow{-7}{*}{ImageNet} &
      \cellcolor[HTML]{E8E8E8}$\text{MMRL}_{\text{ (Ours)}}$ &
      \cellcolor[HTML]{E8E8E8}69.00 &
      \cellcolor[HTML]{E8E8E8}70.30 &
      \cellcolor[HTML]{E8E8E8}\textbf{71.40} &
      \cellcolor[HTML]{E8E8E8}\textbf{72.33} &
      \cellcolor[HTML]{E8E8E8}\textbf{73.40} \\ \midrule
     &
      Linear probe CLIP &
      79.88 &
      89.01 &
      92.05 &
      93.41 &
      95.43 \\
     &
      CoOp &
      92.60 &
      93.07 &
      94.40 &
      94.37 &
      95.57 \\
     &
      CoCoOp &
      93.83 &
      94.82 &
      94.98 &
      95.04 &
      95.16 \\
     &
      MaPLe &
      92.57 &
      93.97 &
      94.43 &
      95.20 &
      96.00 \\
     &
      PromptSRC &
      93.67 &
      94.53 &
      95.27 &
      95.67 &
      96.07 \\
     &
      MMA &
      92.90 &
      94.00 &
      94.33 &
      95.37 &
      96.33 \\
    \multirow{-7}{*}{Caltech101} &
      \cellcolor[HTML]{E8E8E8}$\text{MMRL}_{\text{ (Ours)}}$ &
      \cellcolor[HTML]{E8E8E8}\textbf{94.17} &
      \cellcolor[HTML]{E8E8E8}\textbf{94.83} &
      \cellcolor[HTML]{E8E8E8}\textbf{96.03} &
      \cellcolor[HTML]{E8E8E8}\textbf{96.27} &
      \cellcolor[HTML]{E8E8E8}\textbf{97.13} \\ \midrule
     &
      Linear probe CLIP &
      44.06 &
      58.37 &
      71.17 &
      78.36 &
      85.34 \\
     &
      CoOp &
      90.37 &
      89.80 &
      92.57 &
      91.27 &
      91.87 \\
     &
      CoCoOp &
      91.27 &
      \textbf{92.64} &
      92.81 &
      93.45 &
      93.34 \\
     &
      MaPLe &
      89.10 &
      90.87 &
      91.90 &
      92.57 &
      92.83 \\
     &
      PromptSRC &
      \textbf{92.00} &
      92.50 &
      \textbf{93.43} &
      \textbf{93.50} &
      93.67 \\
     &
      MMA &
      91.23 &
      91.97 &
      92.23 &
      92.77 &
      93.23 \\
    \multirow{-7}{*}{OxfordPets} &
      \cellcolor[HTML]{E8E8E8}$\text{MMRL}_{\text{ (Ours)}}$ &
      \cellcolor[HTML]{E8E8E8}90.87 &
      \cellcolor[HTML]{E8E8E8}91.57 &
      \cellcolor[HTML]{E8E8E8}92.57 &
      \cellcolor[HTML]{E8E8E8}93.03 &
      \cellcolor[HTML]{E8E8E8}\textbf{93.83} \\ \midrule
     &
      Linear probe CLIP &
      35.66 &
      50.28 &
      63.38 &
      73.67 &
      80.44 \\
     &
      CoOp &
      67.43 &
      70.50 &
      74.47 &
      79.30 &
      83.07 \\
     &
      CoCoOp &
      67.22 &
      68.37 &
      69.39 &
      70.44 &
      71.57 \\
     &
      MaPLe &
      66.60 &
      71.60 &
      75.30 &
      79.47 &
      83.57 \\
     &
      PromptSRC &
      \textbf{69.40} &
      \textbf{73.40} &
      77.13 &
      80.97 &
      83.83 \\
     &
      MMA &
      67.87 &
      71.77 &
      76.50 &
      81.40 &
      85.70 \\
    \multirow{-7}{*}{StanfordCars} &
      \cellcolor[HTML]{E8E8E8}$\text{MMRL}_{\text{ (Ours)}}$ &
      \cellcolor[HTML]{E8E8E8}68.70 &
      \cellcolor[HTML]{E8E8E8}72.93 &
      \cellcolor[HTML]{E8E8E8}\textbf{78.17} &
      \cellcolor[HTML]{E8E8E8}\textbf{82.57} &
      \cellcolor[HTML]{E8E8E8}\textbf{86.43} \\ \midrule
     &
      Linear probe CLIP &
      69.74 &
      85.07 &
      92.02 &
      96.10 &
      97.37 \\
     &
      CoOp &
      77.53 &
      87.33 &
      92.17 &
      94.97 &
      97.07 \\
     &
      CoCoOp &
      72.08 &
      75.79 &
      78.40 &
      84.30 &
      87.84 \\
     &
      MaPLe &
      83.30 &
      88.93 &
      92.67 &
      95.80 &
      97.00 \\
     &
      PromptSRC &
      85.93 &
      91.17 &
      93.87 &
      96.27 &
      97.60 \\
     &
      MMA &
      83.60 &
      90.30 &
      93.00 &
      95.97 &
      97.97 \\
    \multirow{-7}{*}{Flowers102} &
      \cellcolor[HTML]{E8E8E8}$\text{MMRL}_{\text{ (Ours)}}$ &
      \cellcolor[HTML]{E8E8E8}\textbf{85.97} &
      \cellcolor[HTML]{E8E8E8}\textbf{91.20} &
      \cellcolor[HTML]{E8E8E8}\textbf{94.60} &
      \cellcolor[HTML]{E8E8E8}\textbf{96.60} &
      \cellcolor[HTML]{E8E8E8}\textbf{98.40} \\ \bottomrule
    \end{tabular}
    }
    }
\end{table*}

\begin{table*}[t]
\small
\centering
\caption{Comparison of MMRL with previous state-of-the-art methods on few-shot learning across 11 datasets.}
\label{few_shot2}
\setlength{\tabcolsep}{15pt}{
\resizebox{0.9\textwidth}{!}{
    \begin{tabular}{@{}ll|ccccc}
    \toprule
    \textbf{Dataset} &
      \textbf{Method} &
      \textbf{1 shot} &
      \textbf{2 shots} &
      \textbf{4 shots} &
      \textbf{8 shots} &
      \textbf{16 shots} \\ \midrule
     &
      Linear probe CLIP &
      43.96 &
      61.51 &
      73.19 &
      79.79 &
      82.90 \\
     &
      CoOp &
      84.33 &
      84.40 &
      84.47 &
      82.67 &
      84.20 \\
     &
      CoCoOp &
      \textbf{85.65} &
      \textbf{86.22} &
      \textbf{86.88} &
      \textbf{86.97} &
      87.25 \\
     &
      MaPLe &
      80.50 &
      81.47 &
      81.77 &
      83.60 &
      85.33 \\
     &
      PromptSRC &
      84.87 &
      85.70 &
      86.17 &
      86.90 &
      \textbf{87.50} \\
     &
      MMA &
      83.03 &
      82.50 &
      82.13 &
      83.00 &
      84.57 \\
    \multirow{-7}{*}{Food101} &
      \cellcolor[HTML]{E8E8E8}$\text{MMRL}_{\text{ (Ours)}}$ &
      \cellcolor[HTML]{E8E8E8}84.87 &
      \cellcolor[HTML]{E8E8E8}85.53 &
      \cellcolor[HTML]{E8E8E8}85.77 &
      \cellcolor[HTML]{E8E8E8}86.33 &
      \cellcolor[HTML]{E8E8E8}87.03 \\ \midrule
     &
      Linear probe CLIP &
      19.61 &
      26.41 &
      32.33 &
      39.35 &
      45.36 \\
     &
      CoOp &
      21.37 &
      26.20 &
      30.83 &
      39.00 &
      43.40 \\
     &
      CoCoOp &
      12.68 &
      15.06 &
      24.79 &
      26.61 &
      31.21 \\
     &
      MaPLe &
      26.73 &
      30.90 &
      34.87 &
      42.00 &
      48.40 \\
     &
      PromptSRC &
      27.67 &
      31.70 &
      37.47 &
      43.27 &
      50.83 \\
     &
      MMA &
      \textbf{28.73} &
      31.90 &
      37.57 &
      44.83 &
      52.70 \\
    \multirow{-7}{*}{FGVCAircraft} &
      \cellcolor[HTML]{E8E8E8}$\text{MMRL}_{\text{ (Ours)}}$ &
      \cellcolor[HTML]{E8E8E8}28.53 &
      \cellcolor[HTML]{E8E8E8}\textbf{34.23} &
      \cellcolor[HTML]{E8E8E8}\textbf{40.47} &
      \cellcolor[HTML]{E8E8E8}\textbf{48.07} &
      \cellcolor[HTML]{E8E8E8}\textbf{57.60} \\ \midrule
     &
      Linear probe CLIP &
      41.58 &
      53.70 &
      63.00 &
      69.08 &
      73.28 \\
     &
      CoOp &
      66.77 &
      66.53 &
      69.97 &
      71.53 &
      74.67 \\
     &
      CoCoOp &
      68.33 &
      69.03 &
      70.21 &
      70.84 &
      72.15 \\
     &
      MaPLe &
      64.77 &
      67.10 &
      70.67 &
      73.23 &
      75.53 \\
     &
      PromptSRC &
      \textbf{69.67} &
      \textbf{71.60} &
      \textbf{74.00} &
      75.73 &
      77.23 \\
     &
      MMA &
      64.00 &
      67.17 &
      69.97 &
      72.30 &
      74.63 \\
    \multirow{-7}{*}{SUN397} &
      \cellcolor[HTML]{E8E8E8}$\text{MMRL}_{\text{ (Ours)}}$ &
      \cellcolor[HTML]{E8E8E8}68.90 &
      \cellcolor[HTML]{E8E8E8}71.53 &
      \cellcolor[HTML]{E8E8E8}73.93 &
      \cellcolor[HTML]{E8E8E8}\textbf{76.00} &
      \cellcolor[HTML]{E8E8E8}\textbf{77.70} \\ \midrule
     &
      Linear probe CLIP &
      34.59 &
      40.76 &
      55.71 &
      63.46 &
      69.96 \\
     &
      CoOp &
      50.23 &
      53.60 &
      58.70 &
      64.77 &
      69.87 \\
     &
      CoCoOp &
      48.54 &
      52.17 &
      55.04 &
      58.89 &
      63.04 \\
     &
      MaPLe &
      52.13 &
      55.50 &
      61.00 &
      66.50 &
      71.33 \\
     &
      PromptSRC &
      56.23 &
      59.97 &
      65.53 &
      69.87 &
      72.73 \\
     &
      MMA &
      52.27 &
      56.90 &
      63.93 &
      67.97 &
      73.47 \\
    \multirow{-7}{*}{DTD} &
      \cellcolor[HTML]{E8E8E8}$\text{MMRL}_{\text{ (Ours)}}$ &
      \cellcolor[HTML]{E8E8E8}\textbf{56.37} &
      \cellcolor[HTML]{E8E8E8}\textbf{61.37} &
      \cellcolor[HTML]{E8E8E8}\textbf{67.87} &
      \cellcolor[HTML]{E8E8E8}\textbf{71.60} &
      \cellcolor[HTML]{E8E8E8}\textbf{75.30} \\ \midrule
     &
      Linear probe CLIP &
      49.23 &
      61.98 &
      77.09 &
      84.43 &
      87.21 \\
     &
      CoOp &
      54.93 &
      65.17 &
      70.80 &
      78.07 &
      84.93 \\
     &
      CoCoOp &
      55.33 &
      46.74 &
      65.56 &
      68.21 &
      73.32 \\
     &
      MaPLe &
      71.80 &
      78.30 &
      84.50 &
      87.73 &
      92.33 \\
     &
      PromptSRC &
      73.13 &
      79.37 &
      86.30 &
      \textbf{88.80} &
      92.43 \\
     &
      MMA &
      55.07 &
      59.80 &
      79.40 &
      86.47 &
      92.37 \\
    \multirow{-7}{*}{EuroSAT} &
      \cellcolor[HTML]{E8E8E8}$\text{MMRL}_{\text{ (Ours)}}$ &
      \cellcolor[HTML]{E8E8E8}\textbf{76.00} &
      \cellcolor[HTML]{E8E8E8}\textbf{82.87} &
      \cellcolor[HTML]{E8E8E8}\textbf{87.67} &
      \cellcolor[HTML]{E8E8E8}88.73 &
      \cellcolor[HTML]{E8E8E8}\textbf{93.37} \\ \midrule
     &
      Linear probe CLIP &
      53.66 &
      65.78 &
      73.28 &
      79.34 &
      82.11 \\
     &
      CoOp &
      71.23 &
      73.43 &
      77.10 &
      80.20 &
      82.23 \\
     &
      CoCoOp &
      70.30 &
      73.51 &
      74.82 &
      77.14 &
      78.14 \\
     &
      MaPLe &
      71.83 &
      74.60 &
      78.47 &
      81.37 &
      85.03 \\
     &
      PromptSRC &
      74.80 &
      \textbf{78.50} &
      81.57 &
      84.30 &
      86.47 \\
     &
      MMA &
      74.17 &
      76.17 &
      80.10 &
      83.43 &
      86.30 \\
    \multirow{-7}{*}{UCF101} &
      \cellcolor[HTML]{E8E8E8}$\text{MMRL}_{\text{ (Ours)}}$ &
      \cellcolor[HTML]{E8E8E8}\textbf{75.97} &
      \cellcolor[HTML]{E8E8E8}\textbf{78.50} &
      \cellcolor[HTML]{E8E8E8}\textbf{82.67} &
      \cellcolor[HTML]{E8E8E8}\textbf{84.67} &
      \cellcolor[HTML]{E8E8E8}\textbf{87.60} \\ \bottomrule
    \end{tabular}
    }
    }
\end{table*}

\end{document}